Paper ID 89

# Vehicle behaviour estimation for abnormal event detection using distributed fiber optic sensing.

**Hemant Prasad[1*], Daisuke Ikefuji[1], Shin Tominaga[1], Hitoshi Sakurai[1], Manabu Otani[2]**

1. NEC Corporation, Japan, h-prasad27@nec.com

2. Central Nippon Expressway Company Limited, Japan

**Abstract**

The distributed fiber-optic sensing (DFOS) system is a cost-effective wide-area traffic monitoring technology that utilizes existing fiber infrastructure to effectively detect traffic congestions. However, detecting single-lane abnormalities, that lead to congestions, is still a challenge. These single-lane abnormalities can be detected by monitoring lane change behaviour of vehicles, performed to avoid congestion along the monitoring section of a road. This paper presents a method to detect single-lane abnormalities by tracking individual vehicle paths and detecting vehicle lane changes along a section of a road. We propose a method to estimate the vehicle position at all time instances and fit a path using clustering techniques. We detect vehicle lane change by monitoring any change in spectral centroid of vehicle vibrations by tracking a reference vehicle along a highway. The evaluation of our proposed method with real traffic data showed 80% accuracy for lane change detection events that represent presence of abnormalities.

**Keywords:**
Vehicle tracking, abnormality detection, lane change detection, distributed fiber-optic sensing

**Introduction**

In an intelligent transportation system (ITS), it is crucial to understand the nature of traffic flow in real-time by employing suitable traffic monitoring systems. Also, there is a need for an effective monitoring system that can provide a cost-effective wide-area traffic flow monitoring solution. However, the widely used traffic monitoring systems, employing traffic cameras and inductive loop detectors, face significant challenge from an economical, monitoring range and time response point of view as they require high installation and maintenance cost due to the need of additional sensors; have a shorter monitoring range; and have slower response time for effectively detecting traffic abnormalities [1][2]. The distributed fiber optic sensing (DFOS) systems that monitor traffic flow using optical fiber cables, that are deployed for communication purposes and already installed along the roads, are a promising technology to solve these issues [3]-[6]. DFOS systems measure vibrations originating from the environment surrounding these fiber cables, such as road traffic, gas

Vehicle behaviour estimation for abnormal event detection using distributed fiber optic sensing.

pipelines, or civil engineering structures, by measuring the change in Rayleigh backscattered signals in the corresponding optical fiber cable section [7][8]. This makes it possible to detect and localize traffic-induced vibrations on roads for effective traffic monitoring [9].

A DFOS system consisting of a sensing device, known as an optical interrogator, connected to a fiber cable deployed along a road section during congestion is shown in Figure 1. The vehicles slow down due to disruption of the traffic flow and the corresponding change in their vibration intensities can be observed in DFOS data as vertical trajectories generated due to these slowing down vehicles. DFOS systems can potentially detect this congestion by monitoring various traffic flow parameters such as speed, count, or vehicle density. To do so, the vehicle trajectories, generated from vibration signals, are recorded on a time-distance plane, also known as waterfall trace. These vehicle trajectories are then used to calculate these traffic parameters and detect traffic flow anomalies for each road section [10]. DFOS systems estimate an average traffic speed by employing algorithms like TrafficNet to denoise DFOS data and extract these vehicle trajectories [11][12]. Therefore, DFOS systems can promptly detect events like the occurrence of congestion.

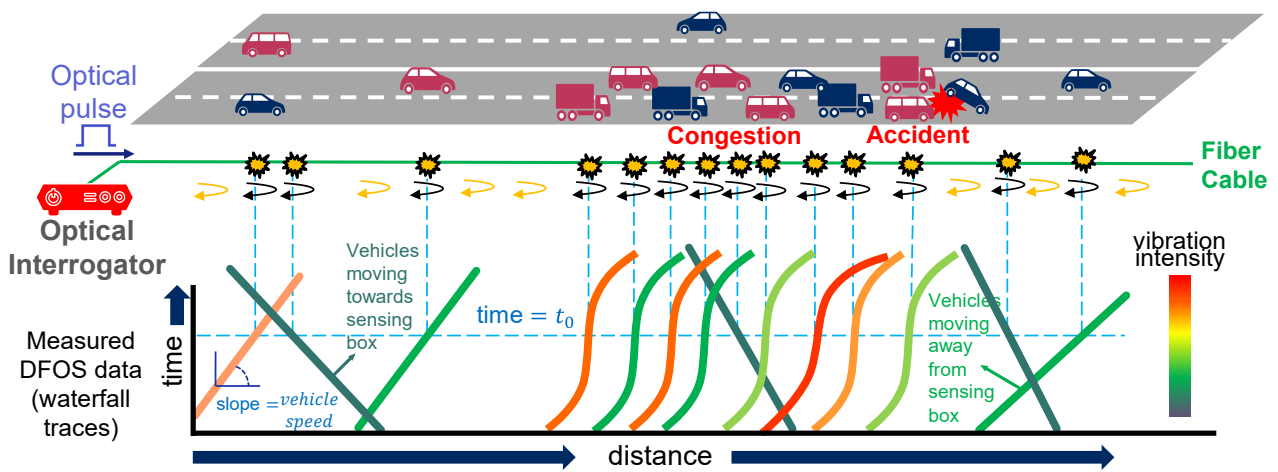

**Figure 1 – A traffic monitoring system using distributed fiber-optic sensing for monitoring traffic flow in one direction on a two-lane road. The upper part of the figure shows schematic diagram of a DFOS system with fiber cable laid along a road during congestion. The lower part of the figure shows corresponding vehicle vibration intensities on time-distance plane (waterfall data). The vehicles travelling on lane closer to the fiber cable (near-lane) produce higher intensities as compared to vehicles travelling on lane farther away from fiber cable (far-lane). Traffic congestions can be observed as vertical vehicle trajectories in the waterfall data.**

In real situations, congestion is often caused due to the presence of abnormalities, such as fallen objects, accidents, or stopped vehicles on any one lane. Here, vehicles tend to change lanes in order to escape these abnormalities or to avoid congestion. However, these vehicles, desiring a lane change to escape abnormalities, can often induce large perturbations in traffic flow resulting in congestion [13][14]. Hence, it is important to identify these lane change behaviors along roads and highways as promptly as possible to detect any abnormalities.

Vehicle behaviour estimation for abnormal event detection using distributed fiber optic sensing.

**Issues for single-lane abnormality detection**

DFOS systems, that employ models like TrafficNet, extract vehicle trajectories and estimate average vehicle speed to monitor traffic flow. However, these systems are incapable of classifying vehicle trajectories based on travelling lane of vehicles because they extract vehicle trajectories irrespective of lane of travel. Figure 2 shows an illustration of DFOS data measured during a single-lane abnormality i.e., a fallen object on near-lane and the vehicle trajectories extracted using TrafficNet for their respective vehicle vibrations. Here, all vehicle trajectories carry equal weight for traffic flow estimation i.e., effect of individual vehicle trajectories based on the lane of travel is not considered. Hence, traffic flow changes occurring on each lane of the road cannot be classified. Therefore, traffic abnormalities due to change in vehicle behaviour on a single lane cannot be detected.

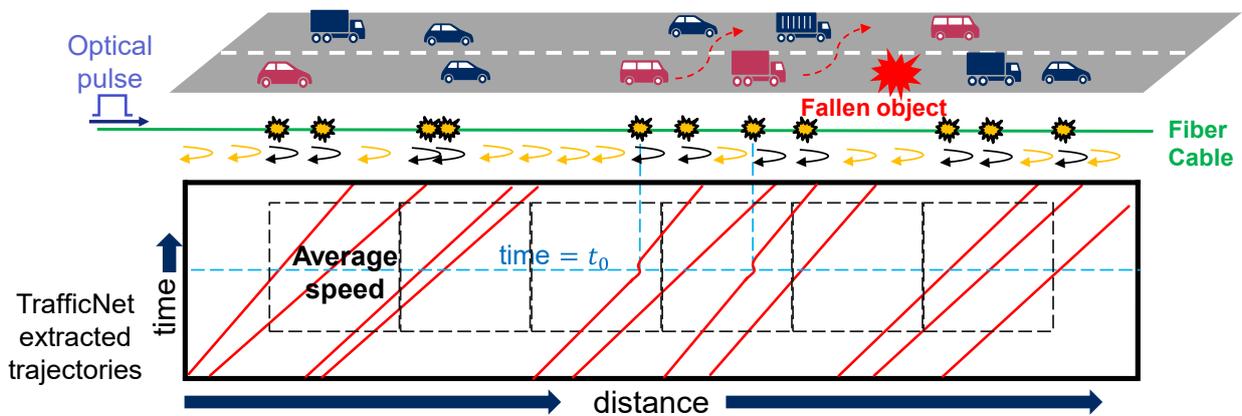

Figure 2 – Waterfall data for distributed fiber-optic sensing system measured during single-lane abnormal event (fallen object). The upper part illustrates the tendency of vehicles to change lanes during occurrence of abnormalities. The lower part illustrates vehicle trajectories extracted from DFOS data using TrafficNet model. TrafficNet cannot distinguish vehicle trajectories based on lane of travel and monitors traffic flow using estimated average traffic speed for all vehicle trajectories.

This leads in difficulties to detect and localize early signs of congestion using conventional DFOS systems where lane of travel is not considered. Here, abnormalities can be detected only when traffic flow for all lanes is disrupted, i.e., abnormalities have already resulted in congestion. Therefore, there is a need to identify smaller changes in traffic flow, due to single-lane abnormalities, in real time. These changes can be detected by monitoring individual vehicle behaviour, such as occurrence of lane changes performed by vehicles avoiding abnormalities.

**Proposed method for single-lane abnormality detection**

To monitor individual vehicle behaviour, such as lane changes, on a road, it is important to identify individual vehicle path and monitor vehicle behaviour. To do so, propose a method to track a vehicle along the road by localizing vehicle positions in DFOS data and checking for any lane change behaviour by monitoring vehicle vibration intensities and frequency changes along the vehicle path.

Vehicle behaviour estimation for abnormal event detection using distributed fiber optic sensing.

The overall block diagram of our proposed method to track individual vehicles and detect abnormalities in single lane is explained in Figure 3. Here, the DFOS data is pre-processed for localization correction and the starting position of vehicle is identified using a camera installed along the road. This start position is then used as a reference to iteratively track vehicle trajectories in DFOS data. Finally, abnormalities due to lane change of this vehicle are detected by monitoring change in vehicle vibration spectral centroid.

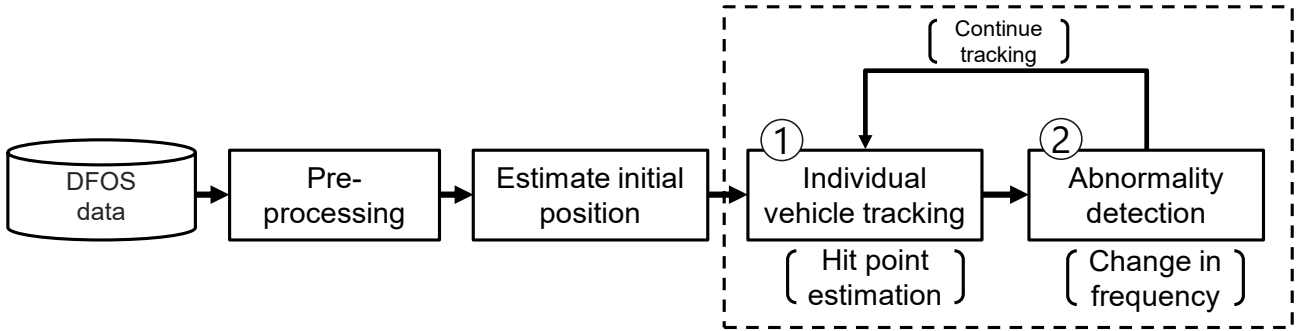

Figure 3 – Overall block diagram of proposed abnormality detection method. Vehicle start position on the DFOS data is located using camera data and a vehicle path for the moving vehicle is estimated using particle tracking. Changes in frequency of vehicle vibration are monitored to detect any abnormalities.

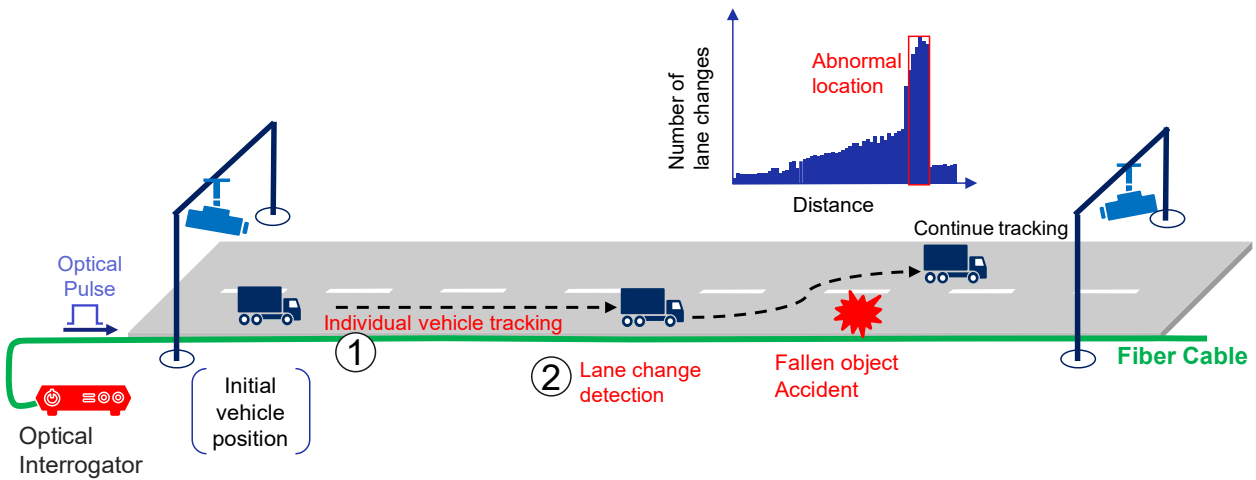

Figure 4 – Illustration of abnormality detection method for a vehicle, travelling on near-lane of a two -lane road, avoiding a fallen object by changing lanes. The initial vehicle position is located using camera information and iterative vehicle tracking is performed. Vehicle behavior is monitored to detect lane changes. The presence of abnormality can be detected by monitoring traffic flow for many lane change events at similar locations.

A step-by-step illustration of our proposed method is explained in Figure 4. Here, a vehicle, travelling on near-lane of a two-lane road and starting at camera position on the right, changes lane from near-lane to far-lane in order to avoid an abnormal event location i.e., a fallen object on the near-lane. Traffic scenarios where multiple vehicles change lanes at the same road section indicate the presence of these abnormalities such as fallen object, traffic accident, or stopped vehicle on that lane.

Vehicle behaviour estimation for abnormal event detection using distributed fiber optic sensing.

*Individual vehicle tracking*

For an in-depth understanding of traffic condition, it is important to understand individual vehicle behaviour on the road. To do so, we propose a method to firstly track individual vehicle path and then detect any abnormalities by identifying lane changes performed by this vehicle on a road. In our proposed method, we perform individual vehicle tracking using an initial position of vehicles estimated with camera installed along the road. We then detect vehicle positions in DFOS data for several time instances by identifying the peak of vehicle vibration generated by these moving vehicles, where these peaks represent the position of vehicles on the road. These peak vibrations, due to a vehicle moving along the fiber cable for any given time instance, are termed as vehicle hit points. The propagation of these vehicle hit points can be observed in DFOS data as the vehicle trajectories or path of the vehicle along the fiber cable. We then construct a path of travel using estimated vehicle hit points for different time instances as explained in Figure 5. The hit points, estimated as part of vehicle trajectories, can then be used to estimate vehicle parameter such as vehicle speed and iteratively continue tracking by calculating the initial position for next set of data.

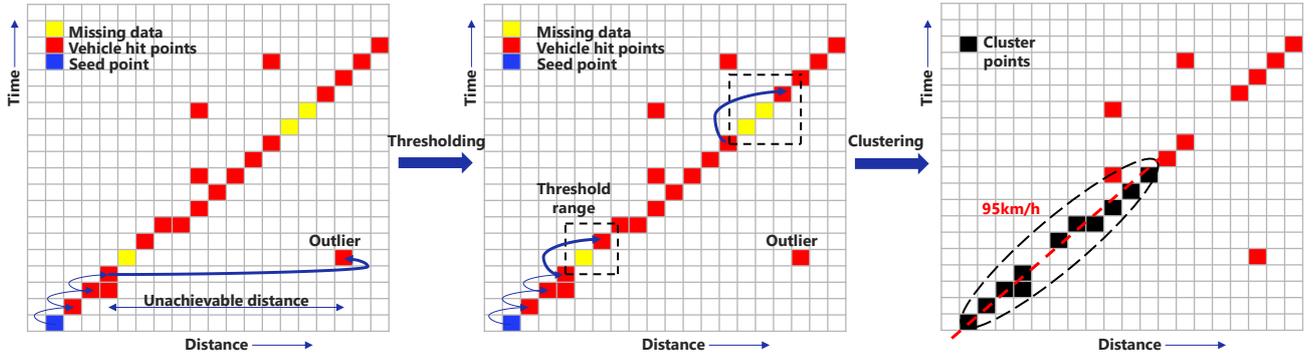

**Figure 5 – Illustration of vehicle hit point de-noising to estimate vehicle path of travel. Hit points closest to the initial position of the vehicle are estimated for each time instance (left). A dynamic threshold range in space-time, based on the data gap and previous speed of the vehicle, is applied (center). The denoised collection of hit points is used to fit a vehicle path using clustering and vehicle parameter is estimated using linear regression (right).**

In our proposed method, we first estimate vehicle hit points for specific time intervals as shown in Figure 5 (right) and then denoise any noisy hit points, arising due to structural or environment noise on roads, by estimating hit points closest to the initial vehicle position. Then, we perform a dynamic thresholding, based on previous vehicle speed and physical limitations, based on distance and time of travel for the moving vehicle, to remove any outliers as shown in Figure 5 (centre). Next, we perform unsupervised clustering of these denoised datapoints using K-Means clustering, chosen for a faster and simpler approach, to fit a vehicle path among these hit points as illustrated in Figure 5 (right) and explained in Equation 1.

$$J = \sum_{i=1}^{m} \sum_{j=1}^{k} w_{ij} \left\| x^i - \mu_j \right\|^2, \qquad (1)$$

where $J$ is the objective function, m is set of vehicle hit points, $k$ is number of vehicles, $x$ is vehicle hit point, $\mu_i$ is mean of points in $S_i$, $w_{ij}$ is '1' for vehicle hit point $x^i$ belonging to cluster $k$; and '0'

Vehicle behaviour estimation for abnormal event detection using distributed fiber optic sensing.

otherwise. The goal is to minimize the objective function w.r.t $\mu_j$ and $w_{ij}$. Here, the datapoints are assigned to initial centroids using Euclidean distance and new cluster centroids are computed for minimised cluster variance. The estimated vehicle path and cluster data points can then be used to estimate the vehicle parameter and behaviour. Here, a least squares method to determine slope of a line of best fit for the cluster datapoints was used to estimate the vehicle speed as explained in Equation 2. This slope was then converted to vehicle speed in kmph using spatial and temporal resolution of DFOS data.

$$m = \frac{\sum_{i=1}^{n}(x_i-\bar{x})(y_i-\bar{y})}{\sum_{i=1}^{n}(x_i-\bar{x})^2} \;, \tag{2}$$

where m is slope of best fit line for cluster datapoints, $\bar{x}$ is the mean of all x-coordinates and $\bar{y}$ is the mean of all y-coordinates.

*Lane change detection*

In this section, a method for detecting lane change behaviour of the tracked vehicles is explained. The measured vibration intensities are inversely proportional to the distance of the source of vibration from the fiber cable. Hence, closer the source, higher the vibration intensities measured by DFOS systems. Due to this, the measured vehicle vibrations vary as the optical fiber cable is usually buried on the shoulder of highways. Furthermore, the frequency characteristics of vehicle vibrations also depend on the lane of travel due to the distance attenuation. Figure 6 shows the frequency characteristics of vibrations, generated by the same vehicle travelling on the near-lane and the far-lane of a road, obtained from real data. Here, we can observe that vehicle vibration frequency is higher for vehicles travelling on lanes closer to the installed fiber cable. Therefore, this difference in frequency characteristics for near-lane and far-lane vehicles can then be used to detect lane change behaviour.

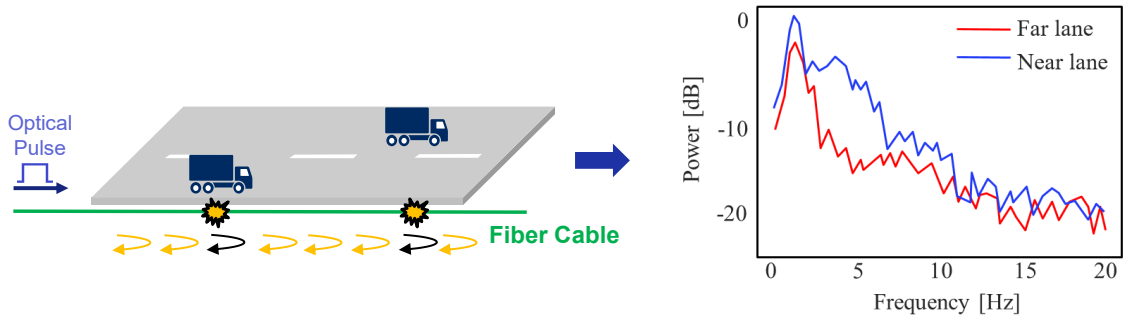

**Figure 6 – Spectrogram of vehicle vibration when vehicle travelling on near-lane (blue) and far-lane (red) on the right. The spectral centroid of vehicles travelling on near-lane are higher as compared to vehicles travelling on far-lane due to distance attenuation of vibrations farther away from fiber cable.**

In this paper, our proposed method employs spectral centroid of vehicle vibration as criteria of frequency characteristics. The spectral centroid shifts lower when a vehicle changes lane from near to far lane. In order to define a threshold for separating near and far lanes, we investigated the distribution of spectral centroids when vehicles drive in each lane. A suitable threshold was calculated, using simulated lane change events for a vehicle, to distinguish near and far lane travel.

Vehicle behaviour estimation for abnormal event detection using distributed fiber optic sensing.

**Evaluation of proposed method**

To evaluate the performance of our proposed abnormality detection method, a field test was conducted at the Shin-Tomei Expressway in the Kanagawa prefecture, Japan. An existing fiber-optic cable infrastructure alongside this highway, that was provided by Central Nippon Expressway Company (C-NEXCO), was used for distributed sensing. The field test was carried out in sections between Isehara-Oyama interchange and Shin-Hadano toll gate. Also, cameras installed at highway locations 8.3KP, 17.17KP, 19.57KP and 20.2KP were used to confirm vehicle positions and to record vehicle travel time. DFOS data of traffic flow was recorded by a DFOS system using the corresponding section of the fiber-optic. We simulated lane change behaviour of vehicles using a test vehicle as reference, and vehicle lane change data was collected using an on-board dash camera installed inside the vehicle. Here, the lane of travel and position of lane change by the test vehicle was recorded. The test vehicle demonstrated multiple lane change behaviours along the Shin-Tomei highway travelling away from the sensing device as shown in Figure 7. Here, a single trajectory on the DFOS data, generated by the test vehicle due to its vehicle vibrations, can be observed.

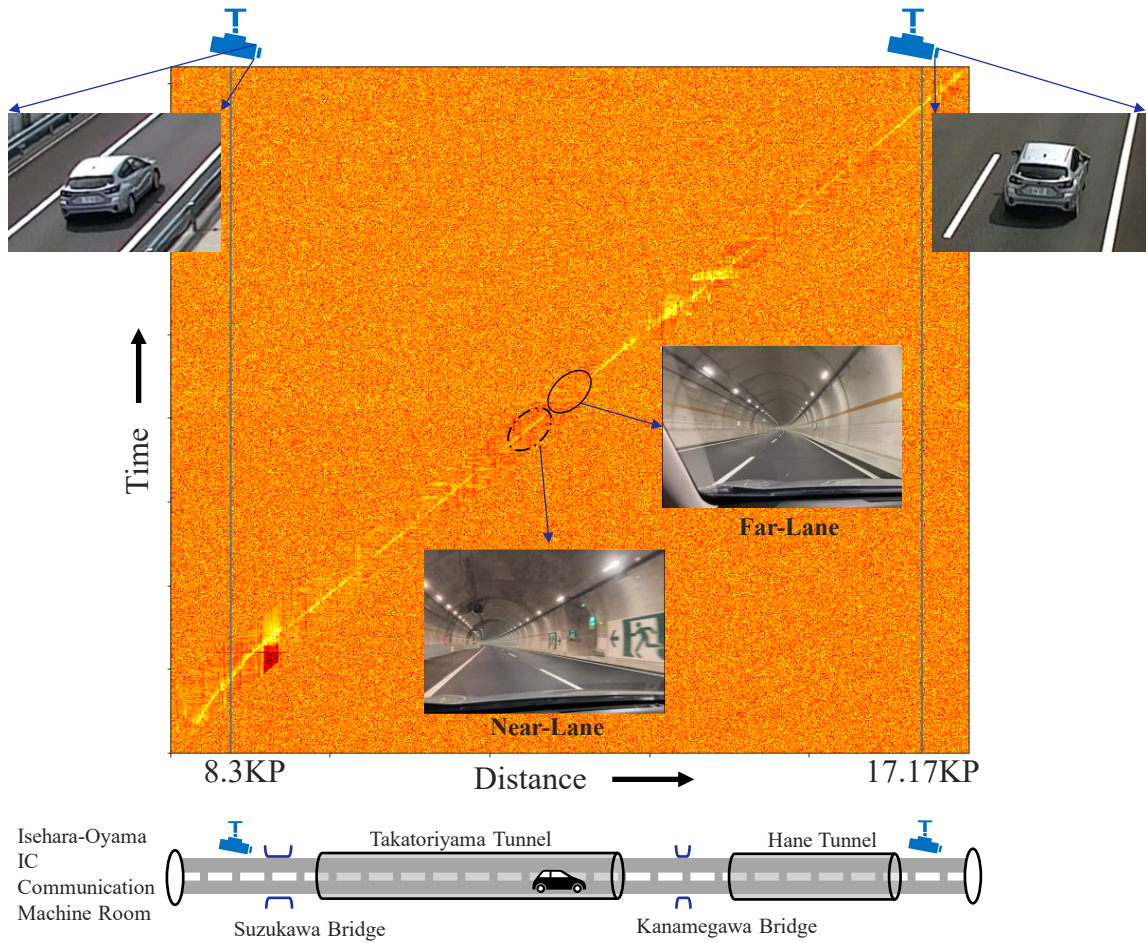

**Figure 7** – Vehicle trajectory of test vehicle travelling on Shin-Tomei highway (above) and corresponding structural information (below). Camera data at 8.3KP to 17.17KP are used to confirm vehicle path and evaluate tracking algorithm performance. The change in vehicle vibration intensity can be observed (center) where near-lane travel produces higher intensities and respective onboard dashcam data is shown.

Vehicle behaviour estimation for abnormal event detection using distributed fiber optic sensing.

The test vehicle is recognized in both camera data i.e., at 8.3KP and 17.17KP on the highway respectively. The time information for vehicle departure from 8.3KP and arrival at 17.17KP is recorded and used to evaluate the performance of tracking algorithm. Here, the change in vehicle vibration intensities for the vehicle trajectory at lane change positions can also be observed at positions highlighted (circled) on the vehicle trajectory in Figure 7 (centre). The vibration intensities are higher for near-lane and lower for far-lane travel and their corresponding dashcam images are shown for reference. The initial position of the vehicle entering the highway near camera location 8.3KP on the highway was used and continuous vehicle tracking until camera position 17.17KP was performed. Here, the vehicle hit points, for a 10sec time interval with new data updated every 1sec, were estimated. Next, we denoised these hit points using shortest path approach with initial position at camera position 8.3KP as reference. Then, we clustered the denoised hit points to estimate vehicle path. These cluster datapoints were used to estimate the vehicle speed and to update the initial position for next cluster. We then iteratively tracked the test vehicle till camera position 17.17KP.

*Individual vehicle tracking analysis*
We performed individual vehicle tracking for our test vehicle travelling away from the sensing device between 8.3KP and 17.17KP locations i.e., 8.87km stretch of Shin-Tomei highway. The entire vehicle path was estimated between both camera locations as shown in Figure 8. We observe a few inaccuracies in path estimation due to bridge damping response, structural noise and fiber installation condition.

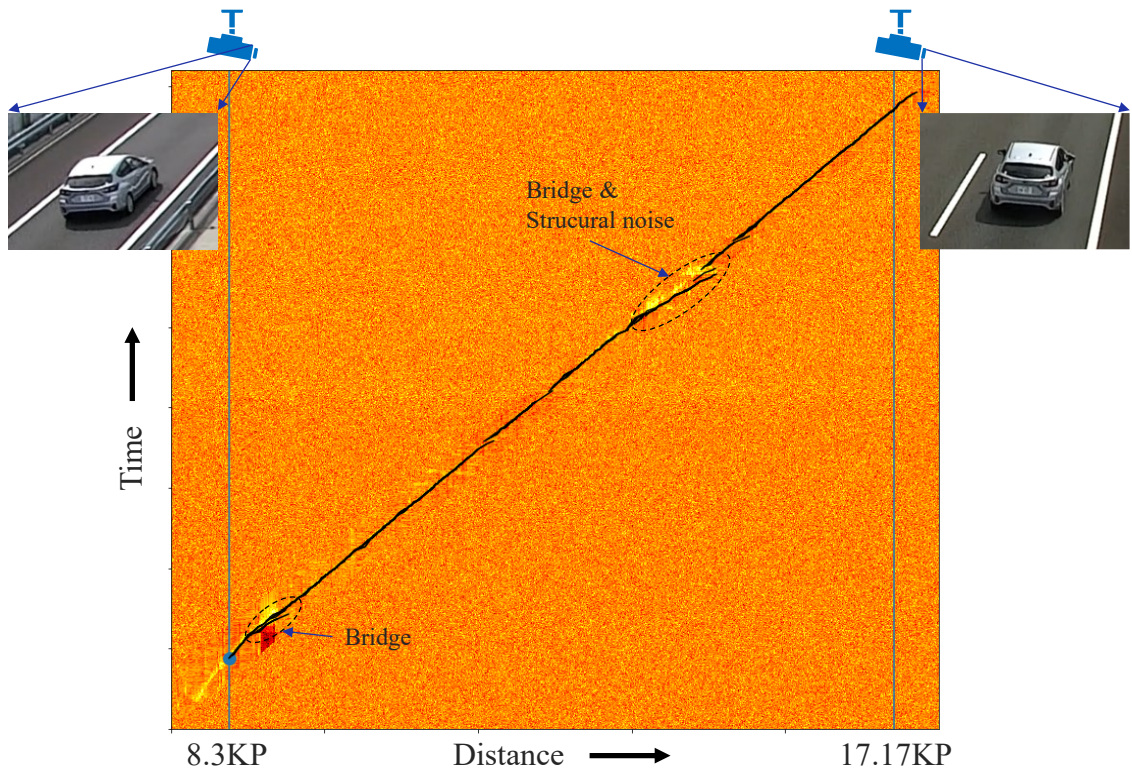

**Figure 8 – Vehicle trajectory of test vehicle (black), travelling on Shin-Tomei highway, estimated using our proposed method for a travel distance of 8.87km between camera positions 8.3KP and 17.17KP.**

Vehicle behaviour estimation for abnormal event detection using distributed fiber optic sensing.

Figure 9 shows the vehicle hit points estimated for vehicle tracking. The initial position of the vehicle, located using a camera, is used to identify the next closest vehicle hit point and remove any noisy hit points among the collection of high intensity points in the DFOS data. This denoised data point is the used for clustering and to estimate vehicle parameter such as speed.

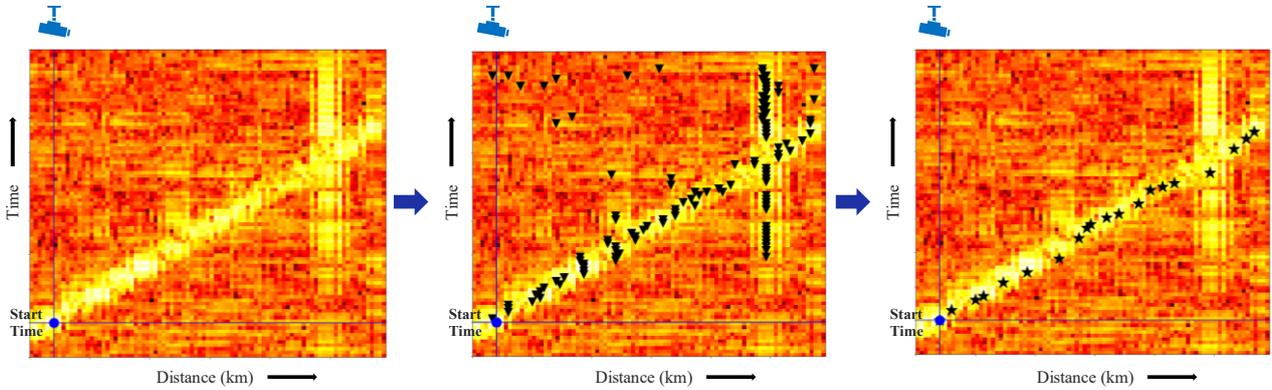

Figure 9 – Estimation of proper vehicle hit points in DFOS data. We consider a vehicle travelling from camera start position shown as a vehicle trajectory travelling from left to right i.e., away from the sensing device (left). Identified peak vibrations positions for all time instances and observe a lot of noisy hit points due to background and structural noise (center). We estimate the closest hit points from initial position and perform thresholding for missing data instances for a collection of denoised hit points of a vehicle path (right).

To analyse the scalability and robustness of our proposed individual vehicle tracking method, we similarly analysed the tracking performance for a stretch of 600m between camera positions 19.57KP and 20.2KP in DFOS data, collected for 1hour time duration, with 124 vehicles. Here, different types of vehicles, such as cars, trucks, tankers, or mini trucks, were considered. The camera data at position 20.2KP was used to identify the initial position of vehicles travelling towards the sensing device for tracking and the camera data at position 19.7KP was used to validate the tracking performance of tracking method by matching the time of arrival from camera data and from vehicle path in DFOS data. If the same vehicle is present in the 19.7KP camera data at respective DFOS estimated vehicle arrival time, proper vehicle tracking is achieved. Similarly, we checked the individual vehicle tracking performance for 124 vehicles travelling from 20.2KP towards 19.57KP of the highway as shown in Table 1. An accuracy of about 87.9% was observed for both large as well as small sized vehicles. The inaccuracies are due to improper vehicle track estimation at noisy sections.

Table 1 – Individual vehicle tracking performance considering type of vehicle.

| Type of vehicle | Vehicles detected | Total Vehicles | Accuracy |
| --- | --- | --- | --- |
| Large vehicles (Truck, van, tanker) | 22 | 25 | 88% |
| Small vehicles (car, minitrucks) | 87 | 99 | 87.87% |
| All vehicles | 109 | 124 | 87.9% |

Vehicle behaviour estimation for abnormal event detection using distributed fiber optic sensing.

*Lane change detection analysis*

To analyse the performance of our lane change approach, we monitored the change in vehicle spectral centroid with all cluster datapoints using vehicle intensities and frequency as parameter. The reference lane change events observed from on-board camera were used to analyse the lane detection accuracy from a total of 30 lane change events between camera positions 8.3KP and 17.17KP for a stretch of 8.87km. Here, the threshold to distinguish lane of travel was estimated to be at 2.7Hz i.e., vehicle vibrations with spectral centroid below 2.7Hz were travelling on the far-lane and vice-versa. The distribution of spectral centroid and threshold frequency for test vehicle vibrations is shown in figure 10.

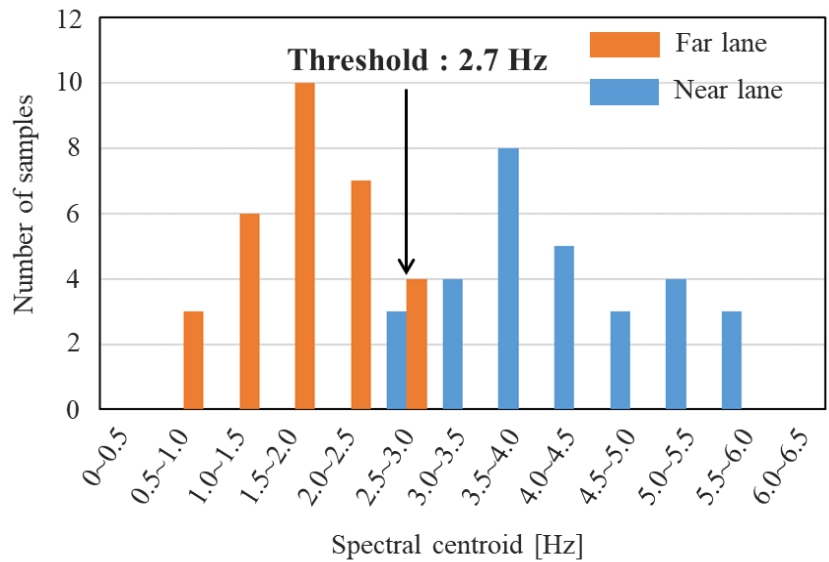

**Figure 10 – Distributions of spectral centroid of vehicle vibration when vehicles drive on near-lane (30 samples) and far-lane (30 samples). The threshold for separating each distribution is calculated at 2.7Hz.**

Table 2 shows the result of evaluation of lane change detection for two modes on lane change i.e., from near to far lane and vice-versa. We confirmed that the proposed method has high performance for lane change detection for vehicle travelling from near to the far lane due to low effect of distance attenuation on near lane. Considering an abnormal event scenario, such as fallen object, demonstrated by multiple lane change events conducted for the test vehicle, this result shows that about 80% of vehicles avoiding fallen objects can be detected. Hence, it is possible to identify abnormal events by detecting many vehicles changing lane at the same location that are avoiding congestion.

**Table 2 – Lane change detection performance considering initial lane of travel.**

| Detection rate | From near to far lane | From far to near lane |
|---|---|---|
| **True positive rate** | 87 % | 77 % |
| **False positive rate** | 13% | 26% |

Vehicle behaviour estimation for abnormal event detection using distributed fiber optic sensing.

**Conclusions**

We presented an abnormality detection method using individual vehicle tracking with distributed fiber-optic sensing data to detect lane change behaviour on roads. The method consists of iteratively tracking a vehicle on a road by first identifying its initial position, estimating vehicle hit points closest to the initial position for several time instances, thresholding any outliers and finally, clustering the hit points to fit a vehicle path. Next, we checked for any abnormalities by monitoring any change in vehicle vibration spectral centroid, which indicates lane change behaviour. We evaluated our proposed method by conducting an experiment with a test car on the Shin-Tomei highway and successfully tracked this vehicle. We conducted lane change experiments and recorded the count and time of lane change; and achieved about 80% accuracy for lane change detection. We also evaluated the performance of our individual vehicle tracking for a distance of 600m between camera positions 19.57KP and 20.2KP as reference and achieved 87.9% accuracy for individual vehicle tracking approach. The inaccuracies in vehicle tracking and lane change detection were caused due to the distance attenuation effect for vehicle moving from far lane to near lane of the highway; and fiber installation conditions that lead to low signal-to-noise (SNR) sections such as bridges, overcrowded and irregular fiber cable installations sections. It is possible to improve these accuracies by processing DFOS data to improve SNR at these sections of road. Thus, identifying lane change behaviour of vehicles can enable us to detect the presence of abnormalities on road. An in-depth understanding of traffic conditions on the road can be determined using our proposed individual vehicle tracking method. The proposed method can have significant impact on time response for identifying early signs of congestion due to abnormalities such as accidents, fallen object, or stopped vehicle.